\documentclass[letterpaper, 10 pt, conference]{ieeeconf}  

\IEEEoverridecommandlockouts                              

\newcommand{\algabbr}{Blox-Net\xspace}

\usepackage{pifont}
\newcommand{\cmark}{\ding{51}}%
\newcommand{\xmark}{\ding{55}}%

\usepackage{float}
\usepackage{graphics}
\usepackage[pdftex]{graphicx}
\usepackage{wrapfig}
\usepackage{enumerate}   
\usepackage{adjustbox}
\usepackage{tabularx}
\usepackage{array}
\usepackage[utf8]{inputenc}
\DeclareGraphicsExtensions{.pdf,.png,.jpg}
\usepackage{epsfig}
\usepackage[font={small}]{caption}
\usepackage{subcaption}
\usepackage[rightcaption]{sidecap}
\usepackage{pbox}
\usepackage{bigstrut}
\usepackage{makecell}
\usepackage{ctable} 

\usepackage{amssymb,amsmath}
\usepackage{gensymb} 
\usepackage{nicefrac}       
\numberwithin{equation}{section} 
\usepackage{algorithm}
\usepackage{algpseudocode}

\usepackage{textcomp} 

\usepackage{array} 
\usepackage{tabularx}
\usepackage{multirow}
\usepackage{multicol}
\usepackage{booktabs}
\usepackage{tabulary}

\usepackage[utf8]{inputenc}
\usepackage{units}
\usepackage{bm}
\usepackage{xspace}
\usepackage{flushend}
\usepackage{balance} 
\usepackage{csquotes}
\usepackage{makeidx}
\usepackage{blindtext}
\usepackage{xcolor}
\usepackage{caption}
\usepackage{subcaption}

\usepackage{url}

\usepackage{hyperref}
\hypersetup{
    colorlinks=true,
}

\usepackage[capitalise, nameinlink]{cleveref}

\usepackage[backend=biber,
            url=false,
            isbn=false,
            doi=false,
            backref=false,
            style=ieee,
            natbib=true,
            mincitenames=1,
            maxcitenames=1,
            citestyle=numeric-comp,
            sorting=none,
            block=none]{biblatex}
\renewcommand{\bibfont}{\small}
\renewcommand{\baselinestretch}{1.0}



\usepackage{expl3}
\ExplSyntaxOn
\newcommand\latinabbrev[1]{
  \peek_meaning:NTF . {
    #1\@}%
  { \peek_catcode:NTF a {
      #1.\@ }%
    {#1.\@}}}
\ExplSyntaxOff
\def\eg{\latinabbrev{e.g}}

\def \MethodAcronym {Blox-Net}

\title{\LARGE \bf
Blox-Net: Generative Design-for-Robot-Assembly 
Using VLM Supervision, Physics Simulation, and a Robot with Reset
}

\author{Andrew Goldberg$^1$, Kavish Kondap$^1$, Tianshuang Qiu$^1$, Zehan Ma$^1$, Letian Fu$^1$ \\Justin Kerr$^1$, Huang Huang$^1$, Kaiyuan Chen$^1$, Kuan Fang$^2$, Ken Goldberg$^1$
\vspace{0.4cm} \\ 
\url{https://bloxnet.org/}
\vspace{-0.2cm} 
\thanks{$^{1}$The AUTOLab at UC Berkeley, $^2$Cornell University}
}

\begin{document}

\maketitle
\thispagestyle{empty}
\pagestyle{empty}

\begin{abstract}
Generative AI systems have shown impressive capabilities in creating text, code, and images. Inspired by the rich history of research in industrial ``Design for Assembly”, we introduce a novel problem: Generative Design-for-Robot-Assembly (GDfRA). The task is to generate an assembly based on a natural language prompt (e.g., “giraffe”) and an image of available physical components, such as 3D-printed blocks. The output is an assembly, a spatial arrangement of these components, and instructions for a robot to build this assembly. The output must 1) resemble the requested object and 2) be reliably assembled by a 6 DoF robot arm with a suction gripper. We then present \algabbr, a GDfRA system that combines generative vision language models with well-established methods in computer vision, simulation, perturbation analysis, motion planning, and physical robot experimentation to solve a class of GDfRA problems with minimal human supervision. 
Blox-Net achieved a Top-1 accuracy of 63.5\% in the ``recognizability" of its designed assemblies (eg, resembling giraffe as judged by a VLM). These designs, after automated perturbation redesign, were reliably assembled by a robot, achieving near-perfect success across 10 consecutive assembly iterations with human intervention only during reset prior to assembly. Surprisingly, this entire design process from textual word (“giraffe”) to reliable physical assembly is performed with zero human intervention.

\end{abstract}

\section{Introduction}


Design-for-Assembly (DfA) has a long history dating back to the start of the Industrial Revolution, where guns, pocket watches, and clocks
were designed with interchangeable parts to facilitate mass production on human assembly lines~\cite{foner2013give}. 
With the advent of industrial automation in the latter half of the 20th century, DfA was expanded to take into account the error tolerances of mechanical assembly systems driven by mechanical cams and belts, and later for robotic assembly systems, the latter known as Design-for-Robot-Assembly (DfRA)~\cite{boothroyd2005assembly}. DfRA is the process of designing a product and robot assembly system together to ensure feasibility, for example designing an injection molded part along with a custom workcell for manipulating it. These design systems were enhanced by the emergence of Computer-Aided Design (CAD) and Computer-Aided-Manufacturing (CAM) software that streamlined human visualization and evaluation of components and assemblies using Finite Element Methods (FEM) and perturbation analysis~\cite{kennard1969computer, chang1991computer, millheim1978bottom, lee1999perturbation}. Such systems help human designers visualize and arrange mechanical components with realistic tolerances, checking for potential jamming and wedging conditions (tolerance stack-up)~\cite{bi2020computer}.

All existing DfRA systems require human designers in the loop~\cite{kennard1969computer, chang1991computer, bi2020computer}. One factor that is difficult for DfRA systems to accurately model is the reliability of robot assembly, which depends on the inherent uncertainty in perception, control, and physics (eg, friction)~\cite{nof1999handbook, goldberg1993orienting, apolinarska2021robotic, tian2022assemble, luo2021learning, fu2023safe}. This can to some degree be modeled with simulation, but it is well-known that 3D simulation systems struggle to accurately model minute 3D deformations and collisions that occur during robot grasping and effects such as deformations of robot gripper and suction cups which can produce substantial errors leading to assembly failures~\cite{xu20206dfc, kim2022ipc, murray2017mathematical, huang2022mechanical, mahler2018dex}. Therefore, physical assembly trials are ideal for evaluation.

\begin{figure}[t]
    \centering
    \includegraphics[width=\linewidth]{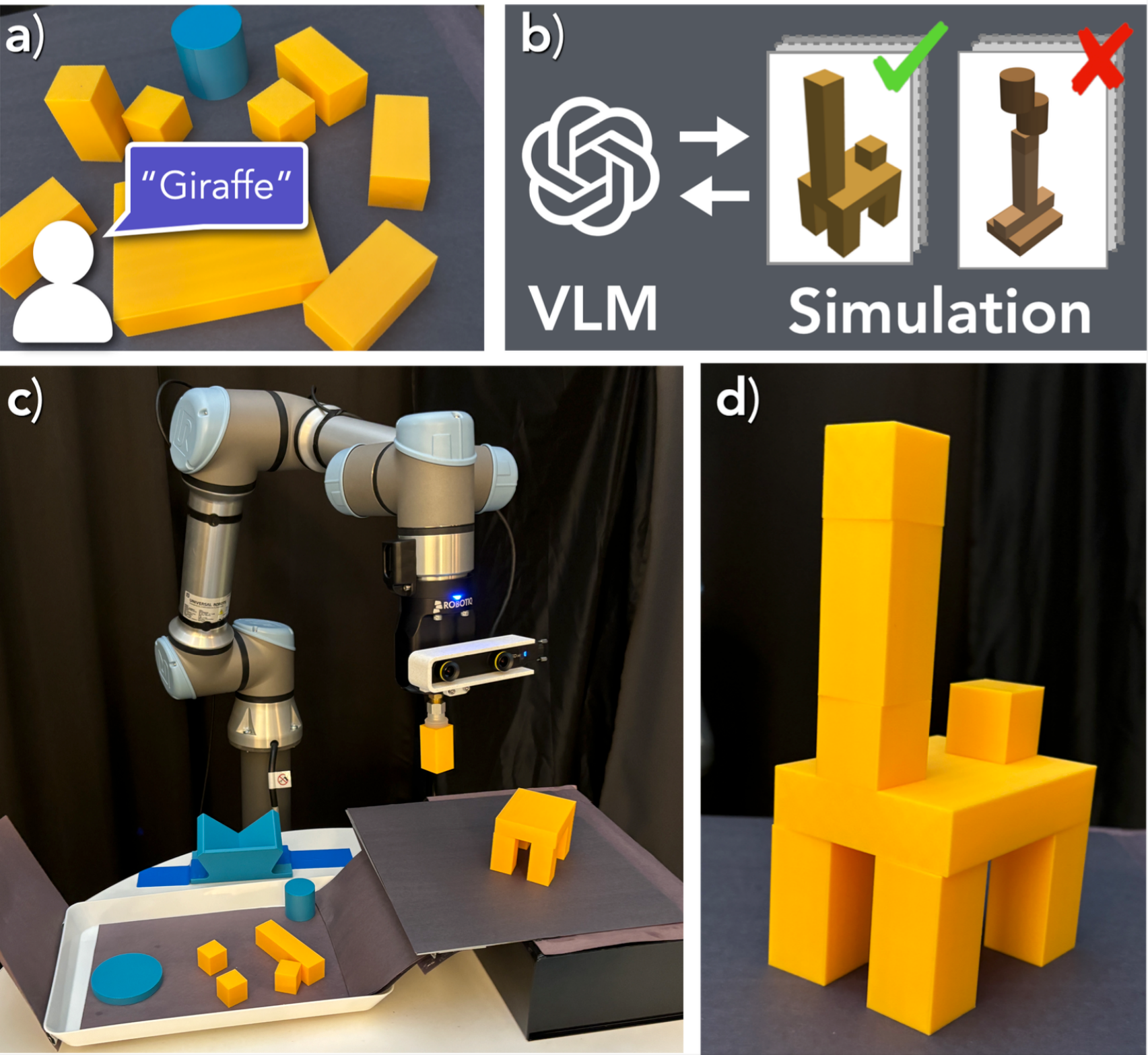}
    \caption{\textit{Can a vision-language model generate designs suitable for robot assembly?} \textbf{\algabbr} is a GDfRA system that produces 3D designs constructible by robots subject to physical material constraints. \textbf{(a)} Starting with a phrase (e.g., "giraffe") and a set of blocks, \textbf{(b)} \algabbr iteratively prompts GPT-4o to generate designs, using simulation to verify stability. \textbf{(c)} A physical robot then assembles the design to test stability and constructibility, \textbf{(d)} resulting in the successful assembly of the design.}
    \label{fig:intro}
\end{figure}

\begin{figure*}[ht!]
    \centering
    \includegraphics[width=\textwidth]{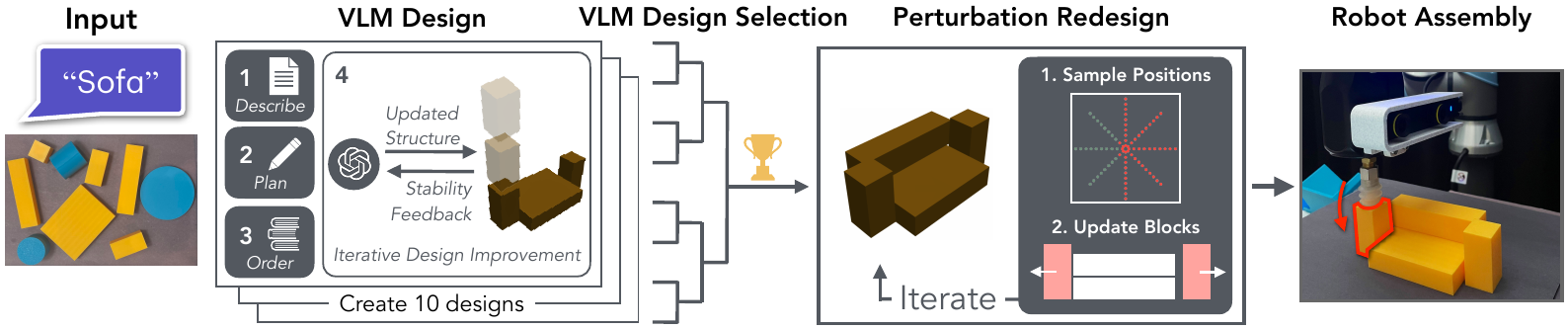}
    \caption{\textbf{Overview of \MethodAcronym.} We present a multi-stage framework for producing physically constructible models based on a user-specified prompt. The Blox-Net pipeline begins with a natural language input and JSON detailing the available blocks. These parameters are passed into a series of VLM prompts, beginning with a high-level overview (\textit{Describe}), followed by requesting specific blocks to use in construction (\textit{Plan}) and a sequence to place them in (\textit{Order}). Finally, the VLM generates the initial design and enters a feedback loop, continuously receiving visual and stability feedback from the simulator. After generating 10 candidate designs, a separate VLM selects the best structure through head-to-head image comparisons. The perturbation redesign phase then adjusts the selected structure to enhance its physical constructability before it is assembled by a real robot.
    }
    \label{fig:method}
\end{figure*}

Recent advances in Generative AI systems have demonstrated remarkable abilities to create novel texts, code, and images~\cite{ramesh2021zero, brown2020language, ouyang2022training}.  Researchers are actively exploring ``text-to-video”~\cite{ho2022imagen, brooks2022generating, castrejon2019improved}
and ``text-to-3D”~\cite{poole2022dreamfusion, lin2023magic3d, wang2023score} systems, where the latter generates 3D mesh structures from textual descriptions (and there are ongoing research efforts applying Gen AI for eCAD design of chips~\cite{liu2023chipnemo}).  This suggests that Generative AI may have potential for DfRA, and that if coupled with a physical robot, it may be possible in certain cases to fully automate the design cycle. 

In this paper, we propose Blox-Net, a fully-implemented \textit{generative} DfRA (GDfRA) system that combines the semantic and text generation capabilities of large language models (LLM) with physical analysis from a simulator. 
Blox-Net includes 3 phases: 1) A vision language model (VLM) with customized iterative prompting to design a feasible 3D arrangement of the available components – an assembly – that approximates the shape of the desired object (eg “a giraffe”);  2) simulation with perturbation analysis to evaluate this assembly in terms of physical robot constructability and to revise the assembly accordingly; 3) Computer vision, motion planning, and control of a physical robot with a camera to repeatedly, through an automated reset, construct this assembly with the given components to automatically evaluate physical assembly reliability. 


This paper makes the following contributions:
\begin{enumerate}
    \item Formulation of a novel problem, Generative Design-for-Robot-Assembly (GDfRA).
    \item Blox-Net, a GDfRA system that combines prompting of GPT-4o with a physical robot, physics simulation, and motion planning to automatically address a class of GDfRA problems where the components are 3D printed blocks.
    \item  Results from experiments suggesting that Blox-Net can produce assemblies – arrangements of given physical blocks – that closely resemble the requested object, are stable under gravity throughout the construction process, and can be reliably assembled by a six-axis robot arm. Starting from singulated objects, Blox-Net achieves 99.2\% accuracy in autonomous block placements.
\end{enumerate}

\begin{figure*}[ht!]
    \centering
    \includegraphics[width=\textwidth]{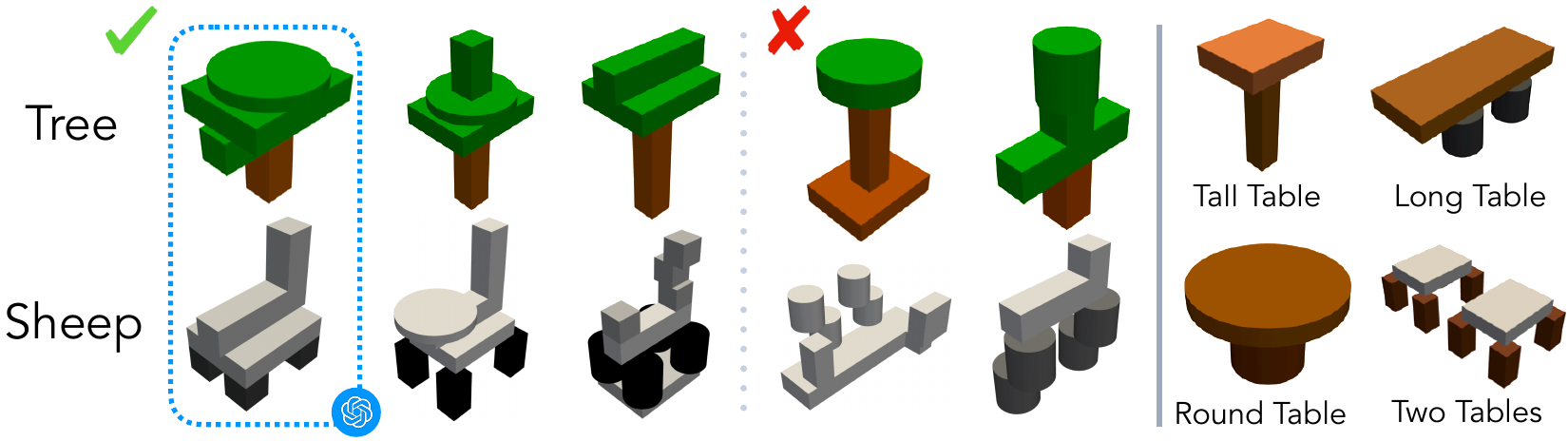}
    \caption{\textbf{Diverse Design Generations:} \textit{Left:} Blox-Net generates a diverse set of candidate designs and uses the VLM (GPT 4o) to select the most suitable one. \textit{Right:} Blox-Net accurately generates a variety of structural designs, adhering to specific input constraints. A set of 10 designs can be generated in 81 seconds, and the selection of the best design takes an additional 60 seconds.
    }
    \label{fig:method}
\end{figure*}

\section{Related Work}\label{sec:related_work}

\subsection{Design for Robot Assembly}


The concept of Design for Assembly (DfA) was pioneered by Geoffrey Boothroyd and Peter Dewhurst in the early 1980s~\cite{boothroyd1983design}, with Hitachi developing its Assemblability Evaluation Method (AEM) in 1986~\cite{miyakawa1986hitachi}. These seminal works laid the foundation for systematic approaches that follow product design guidelines~\cite{ge1960handbook} facilitate facilitate efficient assembly processes. As robotics automation in manufacturing became prevalent, Design for Robot Assembly (DfRA) emerged as an extension of DfA principles, specifically addressing the unique capabilities and limitations of robotic systems in assembly tasks~\cite{Nof1997, boothroyd1987design}.

Design for Robot Assembly (DfRA)~\cite{Nof1997, ohashi2002extended, boothroyd2010product, rampersad1994integrated, rampersad1994integrated} has evolved significantly with the advent of Computer-Aided Design (CAD) and Computer-Aided Manufacturing (CAM) software, which expedite design and evaluation of components and assemblies using Finite Element Methods and perturbation analysis~\cite{kennard1969computer, chang1991computer, millheim1978bottom, lee1999perturbation, bi2020computer}. While these tools facilitate visualization and analysis of tolerances, stresses, and forces, all existing DfRA systems require extensive human input~\cite{kennard1969computer, chang1991computer, bi2020computer}. A persistent challenge in DfRA is accurately modeling assembly reliability, given the inherent uncertainties in perception, control, and physics~\cite{nof1999handbook, goldberg1993orienting, apolinarska2021robotic, tian2022assemble, luo2021learning, fu2023safe}. Simulation can partially address this, but struggles to capture 3D deformations and collisions crucial to robot grasping, necessitating iterative real-world testing and redesign~\cite{tian2023asap, xu20206dfc, kim2022ipc, murray2017mathematical, huang2022mechanical, mahler2018dex}.
Recent advancements leverage large language models (LLMs)~\cite{brown2020language, touvron2023llama} for various aspects of design, including task planning, robot code generation~\cite{macaluso2024toward, you2023robot}, engineering documentation understanding~\cite{doris2024designqa}, and generating planar layouts or CAD models~\cite{gaier2024generative, liu2023chipnemo, badagabettu2024query2cad, wu2023cad}. However, these methods primarily focus on determining assembly sequences for fixed designs. In contrast, this paper addresses both the design and execution aspects of robot assembly, aiming to create physically feasible designs for robotic assembly with minimal human supervision.

\subsection{Text-to-Shape Generation}



Semantic generation of 3D shapes and structures is a long-standing problem in computer vision and computer graphics~\cite{3dmeshgen}. 
Deep generative models have enabled a wide range of approaches that learn to capture the distribution of realistic 3D shapes, in the format of voxel maps~\cite{wu2017learningprobabilisticlatentspace}, meshes~\cite{wang2018pixel2mesh}, point clouds~\cite{fan2017pointset}, sign distance functions~\cite{Park_2019_CVPR}, and implicit representations~\cite{OccupancyNetworks}.
A large number of approaches have also been proposed to reconstruct 3D shapes by conditioning on a single or multiple images~\cite{nerf, yu2021plenoxels, kerbl3Dgaussians, dust3r_cvpr24, liu2023zero}. 
With the advances of aligned text-image representations and vision-language models, an increasing number of works have aimed to generate semantically meaningful shapes specified by natural language instructions~\cite{poole2022dreamfusion, jain2021dreamfields, haque2023instruct}. Unlike these methods, based on the available physical building blocks, Blox-Net generates 3D shapes by prompting an LLM (ChatGPT 4o~\cite{openai2024gpt4osystemcard}) and then generates a plan for assembling the blocks to construct the desired shape.

\subsection{Robot Task Planning with Foundation Models}
Recent advancements in large pre-trained models, such as large language models (LLMs) and vision-language models (VLMs)~\cite{devlin2018bert, radford2018improving, radford2019language, brown2020language, chowdhery2023palm, achiam2023gpt, radford2021learning, li2023blip}, have significantly impacted robotics task planning by leveraging vast internet-scale data. These models enable end-to-end learning through fine-tuning on robotics datasets~\cite{brohan2022rt, brohan2023rt, jiang2023vima, octo_2023, fangandliu2024moka, fu2024icrt} or allow LLMs to directly generate task or motion plans in text or code~\cite{ahn2022can, huang2022inner, huang2022language, chen2023open, liang2023code, singh2023progprompt, wang2023voyager, mirchandani2023largelanguagemodelsgeneral, fangandliu2024moka}. Rather than focusing on motion or waypoint planning, \algabbr prompts the VLM to generate a construction plan by determining the poses of blocks to form semantically meaningful and physically feasible structures, which are then assembled using motion planning and force feedback control.

\section{GDfRA Problem}\label{sec:ps}




We formally define the problem of Generative Design for Robotic Assembly (GDfRA).
We consider the design of a 3D structure that can be assembled with an industrial robot arm (see \Cref{fig:intro}). The input is a word or phrase (\eg, \textit{“bridge”}) and an image of available components for assembly. 
The objective for the GDfRA system is to design a structure which is (1) "recognizable" meaning the structure visually resembles the  provided text input and (2) "constructible" meaning the structure can be assembled by a robot.

\section{Method}\label{sec:method}
We present Blox-Net, a system for a class of GDfRA that assumes (1) components are cuboids and cylinders and (2) components are lying in stable poses within a reachable planar area.

Blox-Net includes three phases. In phase I (\Cref{fig:method}), Blox-Net prompts a VLM (GPT-4o~\cite{openai2024gpt4osystemcard}) to generate multiple assembly designs, from which the VLM selects the top candidate based on stability and visual fidelity. In phase II (\cref{ssec:method_2}), the chosen assembly design undergoes an iterative refinement process in a customized physics simulator. This simulation-based approach applies controlled perturbations to enhance the design's constructability while maintaining its core characteristics.
In phase III (\cref{ssec:method_3}), \algabbr utilizes a robot arm equipped with a wrist-mounted stereo camera and suction gripper to construct the optimized design using 3D printed blocks. The assembly is constructed on a tilt plate, which the robot actuates to automatically reset the blocks back into a tray.

\subsection{Phase I: VLM Design and Selection}
\label{ssec:method_1}



Given the language description and a set of blocks with known sizes and shapes, Blox-Net uses a VLM to generate candidate structure designs. Unlike existing text-to-3D generation methods that produce unconstrained meshes~\cite{jain2021dreamfields, poole2022dreamfusion}, \algabbr generates 3D structures subject to the physical constraints imposed by the available blocks. It prompts the VLM to generate an assembly plan that specifies the 3D locations and orientations for placing each block using the available components (illustrated in \cref{fig:method} (VLM Design Prompting)). 

To facilitate high-quality generation, similar to DALL-E 3~\cite{betker2023improving}, \algabbr first elaborates the prompt. For example, to construct a ``giraffe", the VLM is prompted to give a concise, qualitative textual description that conveys the key features of a giraffe by highlighting the overall structure and proportions. 

After prompt elaboration, the VLM is prompted for the assembly plan. Specifically, the prompt includes the target object (``giraffe"), the VLM's elaboration response, and the set of available blocks. The set of available blocks is encoded as JSON, which provides a structured, flexible format familiar to VLM models. Based on these inputs, the VLM is asked to explain each block's role in the structure.

Once a high-level plan is generated, \algabbr prompts the VLM to produce an assembly plan, specifying the rotation, position and color of objects. Instead of using common rotation parametrizations like Euler angles, \algabbr instructs the VLM to rotate blocks by rearranging their dimensions directly, thereby providing a more simple interface for specifying orientation. Next, \algabbr prompts the VLM to output the (x, y) coordinates for block placement. Limiting the specification to (x, y) coordinates rather than (x, y, z) simplifies the action space and avoids potential issues with blocks being placed inside one another. Blocks are placed by dropping them in the order. 

To enhance stability and correct misplaced blocks, Blox-Net performs iterative, simulation-in-the-loop prompting. Each block's placement is simulated by dropping it in simulation from above the structure. After each placement, the system checks the block for stability. If instability is detected, details such as the specific block that moved, the direction of movement, and two orthographic views highlighting the unstable block are included in a prompt sent back to the VLM for correction. This process continues until all blocks are stable or a maximum of two iterations is reached.

This full prompting pipeline is run in parallel, generating 10 design candidates. For each design, the VLM is queried with a rendered image from the simulation, and provides a rating from 1 to 5 based on how well the structure resembles the intended design. The top-rated stable designs are then paired in a head-to-head comparison, where two images are shown to the VLM, and it selects the more recognizable design. This process is repeated in a knockout format (\cref{fig:method}) until a final design is chosen.

\begin{figure}[t!]
    \centering
    \vspace{3mm}
    \includegraphics[width=\linewidth]{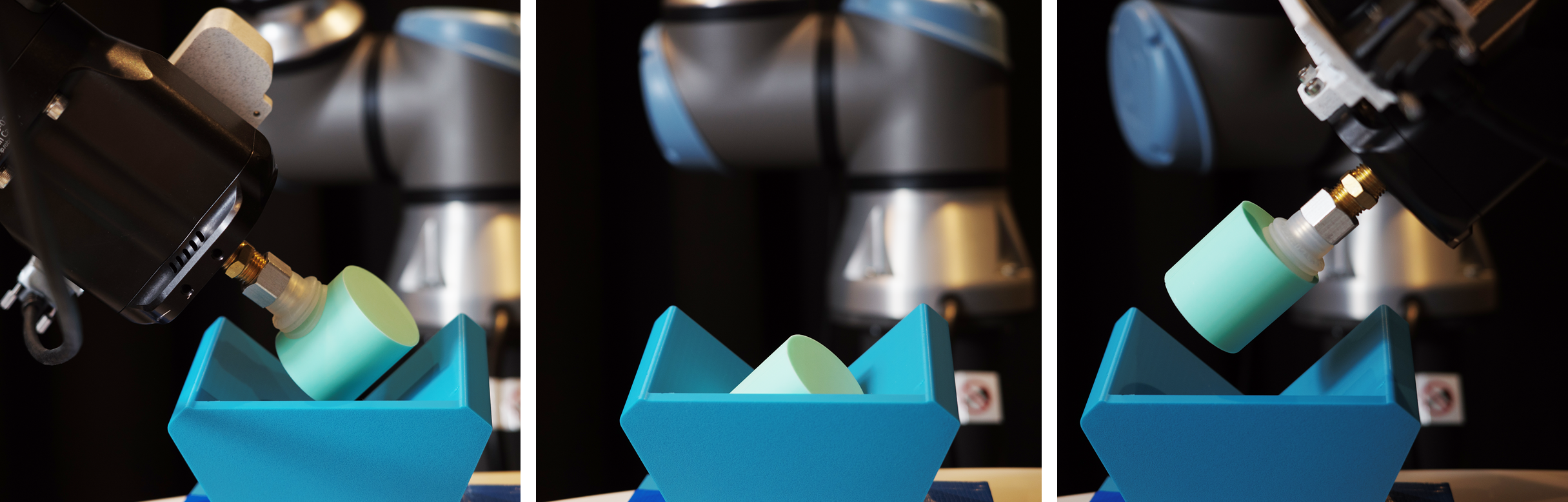}
    \caption{\textbf{Block Reorientation:} The robot first places the block into a 90 degree angle bracket. Then, the block is regrasped on a different face, achieving a 90 degree rotation.}
    \label{fig:reorientation}
\end{figure}

\subsection{Phase II: Perturbation-Based Redesign}
\label{ssec:method_2}

In GDfRA, accounting for imprecise state estimation and robot control is important to ensure robust assembly. The design output from the VLM does not account for such tolerances, which can result in collisions and misplaced blocks during assembly. We thus introduce a perturbation-based redesign process.

The redesign process iterates through each of the blocks and determines if adjustments are needed. A block will be perturbed if it violates at least one of the following three criterion: (1) the surface-to-surface distance to another block is less than a specified collision threshold and the two blocks overlap in the gravity-aligned axis (2) the block is already in collision with another block; or (3) the block is unstable at some nearby sampled point within a predefined radius. 

For each block, Blox-Net samples points evenly along regularly spaced, concentric circles centered at the block nominal location and checks for stability and collision at each point. The block position is updated to the average of positions that are stable and free from collision. This process is applied to all blocks in the structure until no further adjustments are needed or each block has been visited a predefined maximum number of times.

\subsection{Phase III: Robot Assembly and Evaluation}
\label{ssec:method_3}
To evaluate constructability, \algabbr automates physical assembly and evaluates the generated design on a robot. 
The robot first moves to a predefined pose and captures a top-down RGBD image of the blocks on a plastic tray. \algabbr uses SAM~\cite{kirillov2023segany} to segment an RGB image and obtain image masks. SAM segmentations include regions that do not correspond to blocks. To filter out extraneous masks, we generate a point cloud for each mask by deprojecting the masked area from the depth image obtained from a stereo camera~\cite{lipson2021raft}. \algabbr then discards masks that are outside the tray, below a certain minimum area, or not circular or rectangular.

\algabbr refines each mask to segment the top of each block by  fitting a RANSAC~\cite{fischler1981random} plane to the pointcloud and retaining only inliers. The block’s rotation is determined by fitting the tightest oriented bounding box to the refined mask. The block's center is the mean of the points in the filtered point cloud, with the x and y dimensions measured from the point cloud and the z dimension derived from its height relative to the tray base.

Upon determining the size, shape, position, and dimensions for each block, Blox-Net can obtain a new plan through the design generation and perturbation-based redesign process (\cref{ssec:method_1} and \cref{ssec:method_2}), or construct the target object based on a previously generated plan. Blocks may require rotations about their x or y axis to align with the pose used in the plan. This rotation is facilitated by placing the block in a 90-degree angle bracket and regrasping the block from a different side (\cref{fig:reorientation}). After reorientation, the robot captures a new top-down image and all block positions, rotations, shapes, and dimensions are recomputed via the aforementioned pipeline.

The assembly process begins after all blocks are properly oriented. Each block is grasped at its centroid, rotated to the planned orientation, and placed at the location specified in the design. Force feedback control is used for both grasping and placing blocks: during a grasp, the robot lowers onto the block until a force is detected; similarly, during placement, it descends and releases the block once a force is sensed.

To enable efficient testing and design validation, we design an \textit{automatic} reset. After completing the full assembly, the robot arm captures an image. Then, the robot presses down on the tilt plate, dumping the blocks back into the tray. This resets the scene for subsequent trials.
\begin{figure*}[ht!]
    \centering
    \includegraphics[width=\textwidth]{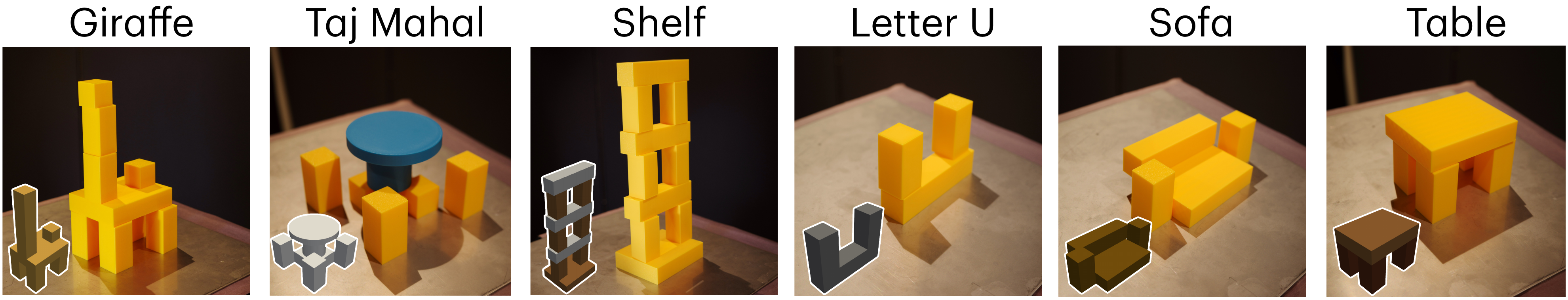}
    \caption{\textbf{Task Execution:} We present Blox-Net VLM generated designs assembled by a robot paired with simulation renderings}
    \label{fig:task_execution}
\end{figure*}

\section{Experiments}\label{sec:experiments}
To evaluate how well the generated structures by \algabbr satisfy the GDfRA objective, we assess both the \textit{semantic recognizability} of the designs (in \cref{ssec:exp_design_1}), which refers to how well the designs semantically align with the prompts, and their \textit{constructability} (in \cref{ssec:exp_design_2}), which refers to how reliably they can be constructed by a real robot. Additionally, we evaluate the effectiveness of the perturbation redesign (in \cref{ssec:perturb_ablation}). To create a candidate objects list for evaluation, we prompt GPT-4o to generate a list of 200 objects spanning categories such as furniture, alphabet letters, architecture, and animals. We run Blox-Net's design generation (\cref{ssec:method_1}) on all objects using a fixed set of block shapes and dimensions. We evaluate semantic recognizability on all 200 designs and evaluate constructability on a representative subset of 11 designs, which showcase the capabilities and limitations of \algabbr, using a physical robot. Additionally, we evaluate the effectiveness of perturbation redesign on 5 designs using a physical robot.


\subsection{Semantic Recognizability}
\label{ssec:exp_design_1}
To measure how well the generated structure resembles the requested language description, we design an experiment using GPT-4o as an evaluator to assess the semantic distinctiveness and accuracy of each design, following methodologies similar to those used in VLM answer scoring~\cite{liu2023llava, vicuna2023, fu2024a}. In this experiment, we use a set of $N$ object labels, where $N$ includes the correct label alongside $N-1$ randomly selected distractor labels from the pool of 200 objects. We provide GPT-4o with a rendered image of the generated assembly and the $N$ labels in random order, and task the VLM with ranking these labels based on how well each one matches the image. We report the percentage of correct Top-1 predictions, and for imperfect guesses, we analyze the average ranking of the correct label within GPT-4o's ordered list, (where a ranking of 1 is best). Additionally, we report the average ranking relative to $N$, with results presented for Top-1 accuracy and average ranking for N=5, 10, 15, 20.


\newcolumntype{P}[1]{>{\centering\arraybackslash}p{#1}}
\newcolumntype{R}[1]{>{\raggedleft\arraybackslash}p{#1}}
\begin{table}[t!]
    \begin{adjustbox}{width=\columnwidth, center}
        \begin{tabular}{R{0.1\linewidth} | P{0.3\linewidth} P{0.2\linewidth} P{0.4\linewidth}}
        \toprule
        
            \textbf{N} & \textbf{Top-1 Accuracy} & \textbf{Avg. Ranking} & \textbf{Relative Ranking} \\
            \midrule
            5  & 63.5\% & 1.7   & 34.0\%   \\
            10 & 48.5\% & 2.92 & 29.1\% \\
            15 & 46.0\% & 3.68  & 24.5\%  \\
            20 & 41.5\% & 5.05  & 25.3\%  \\
            \bottomrule
        \end{tabular}
    \end{adjustbox}
    \caption{\textbf{VLM-Based Design Recognizability}: Top-1 accuracy, average ranking, and relative ranking based on GPT-4o responses averaged across all 200 objects. Relative ranking is reported as the average ranking divided by $N$ where $N$ is the number of labels.
    }
    \label{tab:semantic-recognizability}
    \vspace{4pt}
\end{table}

\subsection{Constructability}
\label{ssec:exp_design_2}
We measure constructability on a real robot over 10 trials on 6 designs selected to highlight diversity. For each trial, we record the \% of blocks correctly positioned at the time of placement, and the \% of trials where the structure is fully successfully assembled. These experiments incorporate automated reset, block reorientation, and assembly fully end-to-end. To assess the system’s autonomy, we track the average percentage of blocks per trial which require intervention during the reset phase, where an intervention is counted for each block moved.
In failure cases after reset where blocks occlude each other, are not adequately separated, or fall out of the tray, blocks are repositioned and placed back in their same stable pose. In cases where the robot fails to regrasp a block during reorientation, the block is placed back in the stable pose corresponding to its final stable pose in the structure. Human interventions are \textit{only} performed during resetting; there is no human intervention during the assembly process.


\subsection{Perturbation Redesign Ablation}\label{ssec:perturb_ablation}
We evaluate the effect of perturbation redesign (\cref{ssec:method_2}) on construction success.
We conduct experiments on 5 objects, each assembled 10 times with and without perturbation redesign. Each trial begins with all blocks singulated and in their correct stable pose. This isolates the effect of perturbation redesign by eliminating influence from prior assembly states. Assembly is performed fully autonomously. We report the percentage of blocks correctly placed at the time of their placement, the average percentage of blocks in the correct location at the end of each trial, and the percentage of trials where the structure is fully completed.

\subsection{Implementation Details}
Blox-Net is implemented with the following components: GPT-4o, PyBullet, UR5e robot arm, Robotiq suction gripper, Zed Mini Stereo Camera, and 3D printed cuboidal and cylindrical blocks.
We use PyBullet as a simulator and define simulation parameters as a uniform object density of $1000 \, \text{kg}/\text{m}^3$, lateral friction coefficient of $0.5$, spinning friction coefficient of $0.2$, gravity of $-9.81 \,\text{m}/\text{s}^2$. A block is measured as unstable if after $500$ simulation steps at $240\text{Hz}$ the block's position deviates by more than $1$cm or is rotated by more than $.1$ radians from its starting position. 

During perturbation redesign, Blox-Net samples $8$ points from each of $10$ concentric circles with radii from $1$mm to $15$mm and each block is perturbed a maximum of 10 times. Blox-Net filters masks by shape by fitting a minimum area bounding rectangle and minimum bounding circle to each mask. Masks are discarded if their area is less than 80\% of the areas of both bounding shapes.

\begin{table}[t]
\vspace{7pt}
\begin{adjustbox}{width=\columnwidth, center}
\begin{tabular}{l| c | c | c}
\toprule
\textbf{Object} & \textbf{\% of Blocks Adjusted } & \textbf{\% Correct} & \textbf{\% of Assemblies} \\ 
(\textbf{\# of Blocks}) & \textbf{During Reset Phase} & \textbf{Blocks Placed} & \textbf{Completed} \\
\midrule
Filament Roll (3) & 14\% & 100\% & 100\% \\
Giraffe (9)       & 28\% & 100\% & 100\% \\
Lighthouse (7)    & 18\% & 100\% & 100\% \\
Letter-U (3)      & 7\%  & 100\% & 100\% \\
Shelf (10)        & 34\% & 100\% & 100\% \\
Table (5)         & 11\% & 98\%  & 90\% \\
\bottomrule
\end{tabular}
\end{adjustbox}
\caption{ \textbf{Robot Assembly and Reset}: 
The table presents the robot assembly results for six designs, each assembled by the robot over 10 trials following a reset, during which all blocks are singulated and reoriented on a plastic tray. Human intervention occurred only during the reset phase to de-stack, singulate, and reorient blocks.
}


\label{tab:full-pipeline-experiment}
\end{table}



\begin{table}[t!]
\begin{adjustbox}{width=\columnwidth, center}
\begin{tabular}{p{0.2\linewidth}  p{0.12\linewidth} p{0.12\linewidth} p{0.12\linewidth} p{0.12\linewidth} p{0.12\linewidth} p{0.12\linewidth} p{0.12\linewidth} }
\toprule
\textbf{Object}& \multicolumn{2}{c}{\textbf{\% Correct Blocks Placed}} & \multicolumn{2}{c}{\textbf{\% Correct in End State}} & \multicolumn{2}{c}{\textbf{\% Full Completion}} \\
& \xmark & \cmark & \xmark & \cmark & \xmark & \cmark \\
\midrule
Ceiling Fan (7) & 66.7\% & \textbf{100\%} & 91.0\% & \textbf{100\%} & 40\% & \textbf{100\%} \\
Sandbox (5) & 50.0\% & \textbf{96\%} & 50.0\% & \textbf{96\%} & 20\% & \textbf{80\%} \\
Shark (6) & 75.0\% & \textbf{100\%} & 76.7\% & \textbf{100\%} & 40\% & \textbf{100\%} \\
Sofa (4) & 72.5\% & \textbf{100\%} & 72.5\% & \textbf{100\%} & 30\% & \textbf{100\%} \\
Taj Mahal (10) & 71.0\% & \textbf{100\%} & 72.0\% & \textbf{100\%} & 10\% & \textbf{100\%} \\
\midrule
Average & 67.1\% & \textbf{99.2\%} & 72.4\% & \textbf{99.2\%} & 28.0\% & \textbf{96.0\%} \\
\bottomrule
\end{tabular}
\end{adjustbox}
\caption{\textbf{Perturbation Redesign}: Each object is followed by the number of blocks in the design enclosed in parenthesis. We run each design for 10 iterations. \xmark\xspace indicates experiments \textbf{without} perturbation redesign and \cmark\xspace indicates experiments \textbf{with} perturbation redesign. \textbf{\% Correct Blocks Placed}: the ratio of blocks that were placed correctly (determined by a group) to the whole structure. \textbf{\% Correct In End State}: the ratio of blocks that remain in their correct pose at the end (blocks may be knocked down by later placements). \textbf{\% Full Completion}: the ratio of overall success (0 or 1 for the whole structure per run over all 10 runs).  We observe a significant performance decrease across all objects in all metrics when constructing without perturbation redesign, demonstrating its impact on design success.}
\label{tab:perturb-analysis}
\vspace{2pt}
\end{table}

\begin{figure}[t]
    \centering
    \includegraphics[width=0.45\textwidth]{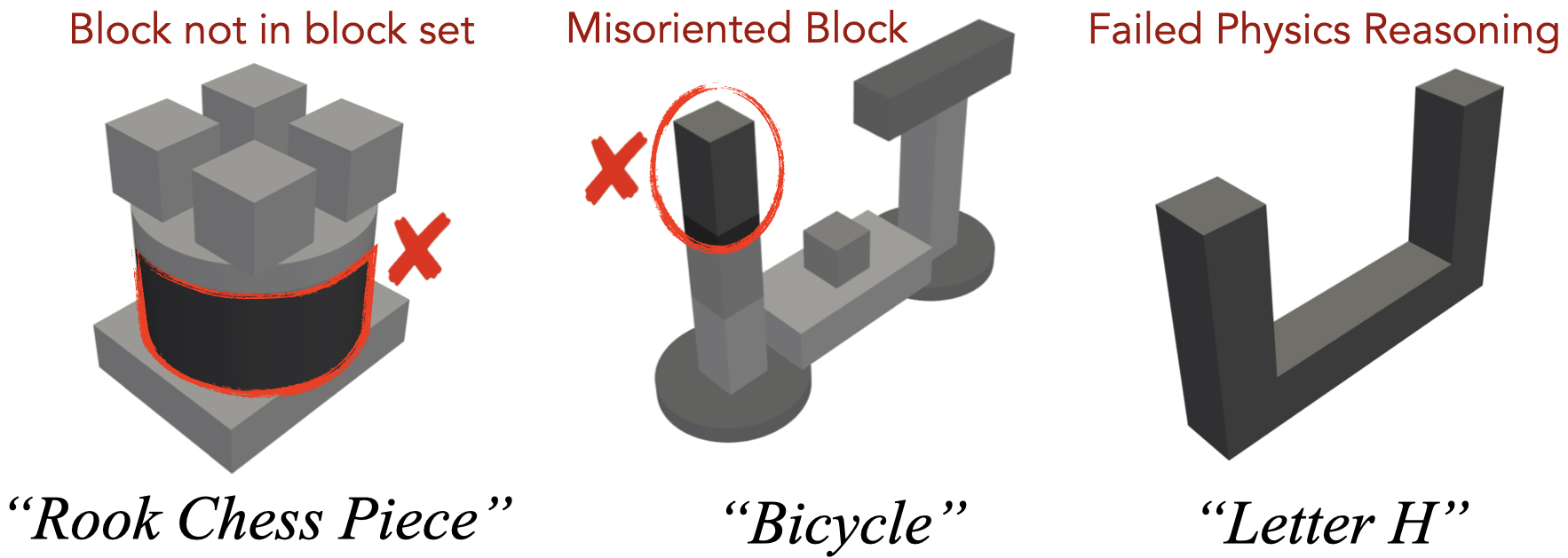}
    \caption{\textbf{VLM Generation Failures} 
    Blox-Net's design generation occasionally produces designs that: include unavailable blocks (Rook Chess Piece), incorrectly orient blocks (Bicycle's Handlebar), or fail to account for gravity (Letter H).
    }
    \label{fig:facilure_analysis}
\end{figure}

\begin{figure}[t]
    \centering
    \includegraphics[width=0.45\textwidth]{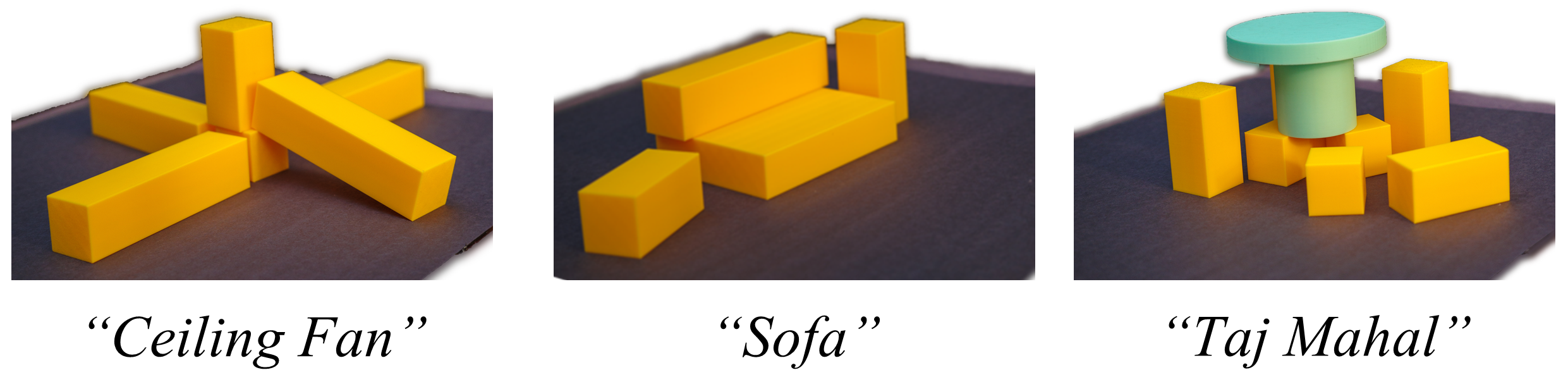}
    \caption{\textbf{Perturbation Redesign Ablation Failures} Omitting perturbation redesign from the Blox-Net leads to a significant increase in physical construction failures. Small inaccuracies in block placement result in collisions, fallen blocks, and structural collapses.
    }
    
    \label{fig:facilure_analysis}
\end{figure}
\section{Results}

\noindent\textbf{Semantic Recognizability}: We present results in \cref{tab:semantic-recognizability}. Results from the evaluation of BloxNet’s designs using GPT-4o as an evaluator suggest that the generated designs closely align with the correct category semantics as recognized by GPT-4o. Notably, with \(N = 5\) labels, the model achieves a Top-1 accuracy of 63.5\%, demonstrating a consistent correspondence between the generated designs and the intended prompts. Importantly, even with larger label sets, the model maintains a reasonable average ranking, with the correct label placed consistently near the top. This suggests that the generated designs remain recognizable, even among a large pool of designs. 

\noindent\textbf{Constructability}: Results are in \cref{tab:full-pipeline-experiment}. All designs are reliably assembled by the robot without human intervention during assembly. Five of six designs achieve a perfect assembly completion rate, and all designs achieve a 98\%+ placement success rate, highlighting Blox-Net's ability to assemble complex structures. Human interventions, which occur only during the reset phase, are sometimes needed to singulate or reorient blocks. 
Complex structures, such as the Giraffe (9 blocks) or shelf (10 blocks), have more human reset interventions due to an increasing likelihood of overlapping, non-singulated, or misoriented blocks.

\noindent\textbf{Perturbation Redesign Ablation} Results are summarized in Table~\ref{tab:perturb-analysis}. Perturbation redesign greatly improves all three metrics across all 5 designs to near-perfect. 
The percentage of correctly placed blocks and the percentage of correctness in the end state are similar for all objects except the ceiling fan. While incorrect block placements typically lead to errors in the final structure, later block placements sometimes correct these errors. Perturbation redesign improves the full completion success rate by an average of 4x. Overall, perturbation redesign significantly enhances the robustness of the assembly process by accommodating slight imprecisions, leading to more reliable and accurate final structures across a variety of designs.

\section{Limitations and Conclusion}

While \algabbr shows promising results in constrained 3D structure generation, it is limited to non-deformable cuboid and cylinder blocks, restricting geometric diversity and reducing Blox-Net's ability to represent complex shapes. Many assembly designs are still not clearly recognizable, likely due in part to these block limitations. The system uses only a suction-based gripper, without accounting for gripper width or slanted surfaces, and sometimes requires human intervention during the reset process, reducing assembly efficacy. 

This paper introduces \algabbr, a novel system addressing the Generative Design-for-Robot-Assembly problem using a three-phase approach: creating the initial designs by prompting a vision language model, conducting simulation-based analysis for constructability, and utilizing a physical robot for assembly evaluation. Experiment results suggest that \algabbr can bridge the gap between abstract design concepts and robot-executable assemblies. Remarkably, five Blox-Net assembly designs, each using 3 to 10 blocks and scoring high in recognizability, were successfully assembled 10 consecutive times by the robot without any human intervention. 


\section*{Acknowledgments}
This research was performed at the AUTOLAB at UC Berkeley in affiliation with the Berkeley AI Research (BAIR) Lab. The authors were supported in part by donations from Toyota Research Institute, Bosch, Google, Siemens, and Autodesk and by equipment grants from PhotoNeo, Nvidia, and Intuitive Surgical. This material is based upon work supported by the National Science Foundation Graduate Research Fellowship Program under Grant No. DGE 2146752. Any opinions, findings, and conclusions or recommendations expressed in this material are those of the author(s) and do not necessarily reflect the views of the National Science Foundation. We thank Timothe Kasriel, Chung Min Kim and Kush Hari for their helpful discussions and feedback.

\renewcommand*{\bibfont}{\footnotesize}
\printbibliography
\clearpage

\end{document}